\documentclass[conference]{IEEEtran}
\IEEEoverridecommandlockouts
\usepackage{cite}
\usepackage{amsmath,amssymb,amsfonts}
\usepackage{algorithm2e}
\usepackage{graphicx}
\usepackage{textcomp}
\usepackage{xcolor}
\usepackage{mathtools}
\def\BibTeX{{\rm B\kern-.05em{\sc i\kern-.025em b}\kern-.08em
    T\kern-.1667em\lower.7ex\hbox{E}\kern-.125emX}}
\usepackage{dcolumn}
\newcolumntype{d}[1]{D{.}{.}{#1}}
    
\begin{document}

\title{Solving the capacitated vehicle routing problem with timing windows using rollouts and MAX-SAT
}

\author{\IEEEauthorblockN{Harshad Khadilkar}
\IEEEauthorblockA{\textit{1. TCS Research  2. IIT Bombay} \\
Mumbai, India \\
harshad.khadilkar@tcs.com, harshadk@iitb.ac.in}
}

\maketitle

\begin{abstract}
The vehicle routing problem is a well known class of NP-hard combinatorial optimisation problems in literature. Traditional solution methods involve either carefully designed heuristics, or time-consuming metaheuristics. Recent work in reinforcement learning has been a promising alternative approach, but has found it difficult to compete with traditional methods in terms of solution quality. This paper proposes a hybrid approach that combines reinforcement learning, policy rollouts, and a satisfiability solver to enable a tunable tradeoff between computation times and solution quality. Results on a popular public data set show that the algorithm is able to produce solutions closer to optimal levels than existing learning based approaches, and with shorter computation times than meta-heuristics. The approach requires minimal design effort and is able to solve unseen problems of arbitrary scale without additional training. Furthermore, the methodology is generalisable to other combinatorial optimisation problems.
\end{abstract}


\section{Introduction} \label{sec:intro}

The vehicle routing problem (VRP) involves the planning of journeys for a fleet of vehicles in order to serve a given set of customer demands. It is closely related to the travelling salesman problem (TSP), with the difference being that the TSP is solved by a single journey while the VRP can be solved by multiple vehicles with exclusive and exhaustive partitioning of the demand nodes. The goal of both TSP and VRP is to minimise the distance travelled (possibly in addition to number of vehicles utilised). A more realistic version of VRP is CVRP-TW, which includes a capacity constraint (C) on the total demand served by any given vehicle as well as time window (TW) constraints attached to each demand node. A precise definition of the problem is given in Section \ref{sec:problem}.

CVRP-TW is highly commercially relevant, with applications in domains such as ride sharing, logistics, supply chain, and various service industries. A detailed overview of solution approaches is given in Section \ref{sec:related}. This paper addresses the following questions which appear to be incompletely answered by present literature:
\begin{itemize}
\item Can CVRP-TW be solved by a less design-intensive method than specialised rule-based systems?
\item Can this method reduce the computational times of meta-heuristic algorithms by utilising offline training?
\item Can the methodology take advantage of known optimisation formulations of CVRP-TW?
\item Can the approach be generic enough to be applied to other combinatorial optimisation problems?
\end{itemize}
The approach described in Sec. \ref{sec:method} resolves these questions through a combination of reinforcement learning to enable policy based rollouts, and a satisfiability solver for optimisation of sub-problems produced by the first step. Experiments are reported on the popular data set shared by Solomon \cite{solomon1987algorithms,solomondata}. While Monte Carlo Tree Search (MCTS) \cite{browne2012survey} offers a way of predicting future outcomes more reliably, the branching factor for realistic problem sizes makes it impractical for real-world use. Neural-MCTS \cite{anthony2017thinking} offers a way out of this dilemma, by allowing us to bias the policy rollouts based on a trained policy or value function. 

\section{Related work} \label{sec:related}

Prior literature in this area is extensive, and can be broadly classified into four categories: heuristic or rule-based systems, meta-heuristics or randomised search, formal optimisation methods, and learning-based methods.

Heuristics have been worked on for many decades \cite{fisher1981generalized,pisinger2007general}. Typically based on specially designed operators such as insertion \cite{campbell2004efficient} and 2-opt \cite{garcia1994parallel,barma20192}. They are the most competitive algorithms for the present problem, with near-optimal results on the Solomon data set \cite{kohl19992,larsen2008recent}. However, there is a lot of design effort involved.

Randomised search algorithms are very popular for vehicle routing. Comprehensive surveys \cite{braysy2005vehicle,gendreau2008metaheuristics} reveal approaches such as Tabu search \cite{chiang1997reactive,backer2000solving}, genetic algorithms \cite{baker2003genetic,hanshar2007dynamic}, and ant colony optimisation \cite{bell2004ant} among many other methods. However, these approaches tend to have long compute times. Prior literature \cite{sultana2021fast,gupta2022deep} shows that the computation times may be more than 100 times longer than those for heuristics or learning based approaches.

CVRP-TW is a combinatorial problem in the NP-hard class, but its formulation using mixed-integer linear programming is well-known \cite{malandraki1992time,kallehauge2008formulations}. There are also variants in literature including robust formulations \cite{agra2013robust} which allow for uncertain travel times as well as online libraries for computing solutions \cite{ortools}. However, the chief challenge in these approaches is the growth in computation times with the scale of the problem, and the unwieldy nature of big-M constraints \cite{lau2003vehicle} required for ordering of customers on routes.

In recent years, the majority of new algorithms use some form of learning approach, either through graph-inspired methods such as pointer networks \cite{vinyals2015pointer,ma2019combinatorial} and graph neural networks \cite{drori2020learning,cappart2021combinatorial} or through reinforcement learning \cite{nazari2018reinforcement, james2019online, delarue2020reinforcement, zhang2020multi}. However, both sets of methods face significant challenges when incorporating constraints. Most studies tend to include vehicle capacity, but they either ignore time windows \cite{nazari2018reinforcement, james2019online, delarue2020reinforcement} or else incorporate them as soft constraints \cite{zhang2020multi}. Furthermore, there do not appear to be any pure learning based approaches that are competitive with existing heuristics in terms of solution quality.

The goal of this paper is to provide a simple composable approach that incorporates the advantages of exact formulations, learning, and meta-heuristics. The algorithm proposed in Sec. \ref{sec:problem} is generic enough to be applied to any combinatorial optimisation problem with a known formulation, and it can also be tuned to trade off the computation time of the solver and search components with the lack of exploration of greedy RL implementations.

\section{Problem description} \label{sec:problem}

The capacitated vehicle routing problem with service time windows (CVRP-TW) assumes that a set of customers $\mathcal{C}$ is known, with their locations $(x_i,y_i)$, demanded load $m_i$, and service time windows $[T_{i,\mathrm{min}}, T_{i,\mathrm{max}}]$, where $i\in\mathcal{C}$, $x_i,\,y_i\in\mathbb{R}$, $T_{i,\mathrm{min}},\,T_{i,\mathrm{max}} \in \mathbb{R}^+$, and $T_{i,\mathrm{min}}< T_{i,\mathrm{max}}$. In this paper, the vehicles start at a single depot $o$, and are able to travel between any two locations (fully connected graph), with the distance $d_{i,j}$ between any two customers $i,j\in\mathcal{C}$ given by the Euclidean metric. The solution approach described in the next section is directly applicable to the multi-depot case as well. We also have a fixed set of vehicles $\mathcal{V}$, the speed $v$ of the vehicles (assumed constant on all edges and for all vehicles in the present work), and the maximum load $M$ that any vehicle can carry in one trip. Then the objective of the problem \cite{sultana2021fast} is to find the total distance $J$ that minimises,
\begin{equation}
    J = \min_{a_{*},\;f_{*},\;l_{*}} \left({\sum_{i,j,k} d_{i,j}\,a_{i,j,k}} + \sum_{i,k} d_{o,i}\,f_{i,k} + \sum_{i,k} d_{o,i}\,l_{i,k} \right),
    \label{eq:objective}
\end{equation}
where $d_{i,j}$ is the distance from customer $i$ to customer $j$, $d_{o,i}$ is the distance from origin (depot) to customer $i$, $a_{i,j,k}$ is an indicator variable which is 1 if vehicle $k$ goes directly from customer $i$ to customer $j$, $f_{i,k}$ is an indicator variable which is 1 if customer $i$ is the first customer served by vehicle $k$, and $l_{i,k}$ is a similar indicator variable which is 1 if customer $i$ is the last customer visited by vehicle $k$. Apart from constraints on $a_{i,j,k}$, $f_{i,k}$, and $l_{i,k}$ to take values from $\{0,1\}$, the other constraints are defined below in brief. A detailed description can be found in \cite{sultana2021fast}.

Every customer must be served within its specified time window. If $t_{i,k}$ is the time at which vehicle $k$ visits $i$, then 
\begin{equation}
T_{i,\min} \leq t_{i,k} \leq T_{i,\max}, \text{ if }f_{i,k}=1\text{ or }\exists j\text{ s.t. }a_{j,i,k}=1 \label{eq:tw}
\end{equation}
The total load served by a vehicle must be at most equal to its capacity $M$. This is formalised as,
\begin{equation}
\sum_{i,j} m_{j}\,a_{i,j,k} + \sum_i m_{i}\,f_{i,k} \leq M \quad \forall k \label{eq:maxload}
\end{equation}
Travel time constraints are applicable between any two locations, based on the distance between them, the speed $v$ at which vehicles can travel, and the pre-specified service time $\Delta_j$ required at each customer.
\begin{align}
    t_{i,k} \geq \frac{d_{o,i}}{v} \text{ if }f_{i,k}=1  \label{eq:depottimestart} \\
    t_{i,k} \geq t_{j,k} + \Delta_j + \frac{d_{j,i}}{v} \text{ if } a_{j,i,k}=1 \label{eq:intercustomer} 
\end{align}
Clearly, this is a simplified version of a real-world situation where the number of vehicles also needs to be minimised, the vehicles can have different distance and velocity constraints, travel times can vary based on traffic conditions, vehicles can do multiple trips, in addition to other variations. The dynamic version of the problem will also allow customers to `pop up' at arbitrary times, while a plan is being executed. However, the formulation described above is itself of interest because of its NP-hard nature, and the fact that exact solutions quickly become intractable as the problem size increases.

\section{Methodology} \label{sec:method}


\subsection{High-level description}

An overview of several decades of work on this problem is given in Sec. \ref{sec:related}. One may generalise the outcomes as follows. Heuristics are the most competitive algorithms currently known in terms of solution quality and computation time, but require significant design effort. Furthermore, they cannot be generalised to other combinatorial optimisation problems. Meta-heuristics are similarly competitive, but have high computation times. Exact optimisation becomes rapidly infeasible as the problem scale increases. Finally, learning based solutions face challenges with respect to solution quality.

Consequently, the methodology proposed in this paper combines three different ideas into a single integrated approach as outlined in Algorithm \ref{alg:pseudo}. The meta-heuristic inspiration is used in the form of stochastic policy rollouts to identify good routing decisions. However instead of randomised local search, the rollouts are performed by a pre-trained reinforcement learning algorithm. The immediate forward path identified by rollout is optimised using a satisfiability solver in MAX-SAT mode. Finally, a post-processing MAX-SAT step further optimises the final routes. The individual components are described below.

\begin{algorithm}[t]
 \KwData{Customer data specification}
 \KwResult{Vehicle routes and service times}
 Initialise: Single vehicle at depot, parameter $\kappa$\;
 \While{at least one customer yet to be served}{
  identify further feasible customer-vehicle pairs\;
  shortlist top $\kappa$ pairs identified by RL\;
  do stochastic rollouts using RL  policy\;
  choose the decision with the lowest total distance\;
  pick sub-tour being served by chosen vehicle\;
  optimise sub-tour using forward SAT\;
  \If{vehicle leaving depot}{
   spawn a new vehicle at depot\;
   }
  implement the optimised sub-tour\;
 }
 Finalise: Optimise vehicle tours with tightening SAT\;\vskip2pt
 \caption{Running an episode of CVRP-TW. The same procedure is used during training and testing, with the exception that exploration steps are taken in a  uniformly random fashion instead of the learnt RL policy.}
 \label{alg:pseudo}
\end{algorithm}

\subsection{Preprocessing of data set} \label{subsec:preproc}

Each data set first undergoes a preprocessing step when it is loaded, in order to characterise its properties. A simple procedure as described in Algorithm \ref{alg:cluster} separates all the customers into clusters, which are used for feature computation in Sec. \ref{subsec:rl}. The algorithm is designed to work without predefining the number of clusters desired, as opposed to the typical k-means or related procedures \cite{likas2003global}. Further characteristic quantities include a time threshold $\tau$ equal to the median travel time between all pairs of customers in the data set, the maximum time window $t_{max}$ in the data set, and the largest inter-customer distance $d_{max}$ in the data set. Each of these quantities is used in feature computation as described below.

\begin{algorithm}[t]
 \KwData{Customer locations, parameter $n\in\mathbb{Z}^+$}
 \KwResult{Clusters, neighbourhood radius $\rho$}
 Compute Euclidean distances $d_{i,j}$ between customers\;
 Initialise set $\mathcal{K}$ of clusters as empty set\;
 \While{at least one customer has no cluster mapping}{
  Define a new empty cluster $\Phi$\;
  Add nearest unmapped customer from depot to $\Phi$\;
  \For{all customers in $\Phi$}{
   Add nearest $n$ neighbours to $\Phi$ if these neighbours have no existing mapping\;
   \If{no new customers got added to $\Phi$}{break\;
   }
   }
   Add cluster $\Phi$ to $\mathcal{K}$\;
 }
 Set neighbourhood radius $\rho$ as half of the largest cluster diameter in $\mathcal{K}$\;
 Finalise: Set of clusters $\mathcal{K}$ and radius $\rho$\;\vskip2pt
 \caption{Cluster preprocessing pseudo-code.}
 \label{alg:cluster}
\end{algorithm}

\subsection{Reinforcement learning for routing} \label{subsec:rl}

At each step in the rollout, each active vehicle-customer pair is evaluated independently, similar to the suggestion in prior work \cite{gupta2022deep, sultana2021fast}. The canonical reinforcement learning (RL) problem is specified by a Markov Decision Process $(S,\mathcal{A},\mathcal{R},T,\gamma)$, with $S$ denoting the states, $\mathcal{A}$ denoting the actions, $\mathcal{R}$ denoting the rewards, $T$ the transition probability $S\times \mathcal{A}\rightarrow S$, and $0\leq \gamma \leq 1$ being a discount factor on future rewards \cite{sutton2018reinforcement}. In the present case, the states are defined with respect to a specified vehicle-customer pair in the context of all the customers in the data set. These features are listed in Table \ref{tab:inputs}. A neural network $Q(S,\mathcal{A}):S\times \mathcal{A}\rightarrow \mathbb{R}$ is used to estimate the \textit{value} \cite{sutton2018reinforcement} of each feasible vehicle-customer pair (as defined by ability of vehicle to arrive at the customer before time window closure with sufficient capacity to meet customer demand).

The reward $\mathcal{R}$ associated with each decision is derived from the sequence of states as observed by each vehicle separately. The transition function $T$ thus applies to trajectories of individual vehicles rather than to the episode as a whole (since different vehicles could move in successive RL steps, it does not make sense to trace the raw sequence of states observed in the episode). Let a vehicle $k$ serves $p=\{1,\ldots,P\}$ customers in its route, with each leg being of length $d_p$ and requiring time $t_p$ between time of service completion at the previous customer and the time of service start at the current customer. Then the reward for each decision is,
\begin{equation}
\mathcal{R}_{k,p} = \frac{\rho-d_p}{d_{max}} + \frac{\tau-t_p}{t_{max}} + \gamma^{P-p}\mathcal{R}_{term}\text{, where}
\label{eq:reward}
\end{equation}
\begin{equation}
\mathcal{R}_{term}=2\,\rho-\frac{1}{P+1}\,\left( \sum_p d_p + D_{return} \right).
\label{eq:terminal}
\end{equation}
The reward defined in (\ref{eq:reward}) provides an incentive for shorter legs in terms of both distance and time. The terminal reward $\mathcal{R}_{term}$ defined in (\ref{eq:terminal}) compares the average leg distance on a vehicle journey (including the final return leg $D_{return}$ to the depot) to the largest cluster diameter $2\,\rho$. The goal of training the neural network is to minimise the mean squared error between its output and the realised reward $\mathcal{R}_{k,p}$ for each decision.

The input is a vector of size 17, as specified in Table \ref{tab:inputs}. This is followed by a fully connected neural network with layers of size (128, 64, 32, 8) and an output layer of size 1 (scalar value). The hidden layers have \texttt{tanh} activation. A memory buffer of maximum $2^{16}=$ 65,536 samples is maintained, with training happening after every 10 steps with a batch of size 4096 samples. The learning rate is 0.001. Exploration is implemented using an $\epsilon-$greedy policy with $\epsilon$ decayed from 1 to 0 with a factor of 0.9995 after each episode. Exploration steps are taken uniformly randomly, while exploitation steps are chosen using a \texttt{softmax} function over the values attached to vehicle-customer pairs.

\begin{table}
\centering
\caption{Inputs for evaluating the value of each vehicle-customer pair, with $loc$ as current location, $c$ as proposed customer, and $v$ as the proposed vehicle. All features are normalised as explained in text.}
\label{tab:inputs}
\begin{tabular}{|p{1cm}|p{6.5cm}|}
\hline
Input & Explanation \\
\hline
$d$ & Distance from $loc$ to $c$ \\
$b_\mathrm{d,short}$ & Is $d$ smaller than neighbourhood radius $\rho$ \\
$t$ & Time gap from now to start of service at $c$ \\
$b_\mathrm{t,short}$ & Is time gap within time threshold $\tau$ \\
$ngb$ & Are $loc$ and $c$ part of the same cluster \\
$non\_d$ & If $ngb=1$, distance from $c$ to nearest non-member \\
$c\_left$ & If $ngb=0$, are any in-cluster customers left unserved \\
$drop_{far}$ & If $c\_left=1$, are the dropped customers farther from depot than $loc$ \\
$drop_{cls}$ & If $c\_left=1$, are dropped customers within a distance of $\rho$ from $loc$ \\
$drop_{long}$ & If $c\_left=1$, is the distance from dropped customers to nearest non-member more than distance from $loc$ to dropped customers \\
$served$ & How many customers of $c$ cluster has $v$ served so far \\
$cls_{dem}$ & Could $v$ serve all cluster members of $c$ based on demand \\
$hops$ & How many cluster members of $c$ can be served before $c$ \\
$cls_{tim}$ & Is every cluster member feasible following $c$ \\
$urgt$ & How close to time window closure of $c$ is $v$ arriving \\
$dfrac$ & Ratio of time being consumed for serving $c$ to the fraction of $v$'s capacity being consumed \\
$remote $& How remote is the neighbourhood of $c$ \\
\hline
\end{tabular}
\end{table}

\subsection{Policy rollouts}

As explained in Sec \ref{sec:related}, the drawback of using vanilla RL is the unpredictable nature of the long-term effects, specifically the difficulty of estimating $\mathcal{R}_{term}$. Carrying out partial or complete rollouts using a trained policy is a popular recent improvement on Monte-Carlo Tree Search, and is called Neural-MCTS \cite{anthony2017thinking}. In this paper, the rollout is carried out up to episode completion for the top $\kappa$ vehicle-customer pairs, as explained in Algorithm \ref{alg:pseudo}. At each time step, the learnt RL policy chooses a vehicle-customer pair based on a \texttt{softmax} function over the estimated values. Only feasible pairs are included, based on the ability of a given vehicle to serve a proposed customer within the time window and with sufficient demand, as defined by (\ref{eq:maxload}) and (\ref{eq:intercustomer}). 

Clearly, there are several other ways of neural MCTS such as partial episodic rollout followed by estimation of residual value. One may also experiment with different values of $\kappa$, which is described in Sec. \ref{sec:results}. Finally, one may perform multiple rollouts in order to maximise the probability of finding good solutions, along similar lines as most meta-heuristics. A specific choice about the breadth, depth, and repetitions of rollout can be made based on the problem at hand and the available computational resources.

\subsection{Satisfiability solver formulation}

The solver chosen for optimisation of sub-problems in this work is Z3 \cite{moura2008z3}. Given the underlying nonlinear nature of the problem as described in (\ref{eq:objective})-(\ref{eq:intercustomer}), the typical mixed-integer linear program for CVRP-TW uses big-M constraints \cite{raman1994modelling} to handle the conditional statements. However, this approach has well-known computational challenges. A satisfiability (SAT) solver, on the other hand, provides a way of naturally specifying logical constraints, and modern SAT solvers are known to be both efficient and able to handle optimisation objectives through the `MAX-SAT' approach \cite{narodytska2014maximum}.

Experiments with different ways of formulating the MAX-SAT problem for CVRP-TW showed that Z3 could handle strongly nonlinear constraints, but struggled with a larger number of decision variables. This appears to be in clear contrast with mixed-integer program solvers, where simplicity of constraints is prioritised over the number of variables. Based on this intuition, I eliminated the connectivity variables $a_{j,i,k}$ (even with a fixed vehicle $k$, the number of variables increases as $|\mathcal{C}|^2$) as well as the first/last indicators $f_{i,k}$ and $l_{i,k}$. In lieu of these variables, we shall specify the route of the vehicle using an integer order variable $O_{i,k}\in\{1,2,\ldots,|\mathcal{C}|\}$.

We also take inspiration from work on constrained position shifting (CPS) from air traffic management \cite{balakrishnan2006scheduling} to improve the search efficiency of the SAT solver. Intuitively, CPS allows for a maximum deviation in the planned order (compared to the route generated by rollout) up to a maximum limit of $\Delta$. This parameter is successively tightened until an optimal solution is produced in the set time limit. The problem with $\Delta=0$ has a trivial solution identical to the rollout route, although opportunistic customers may be added to this as explained below. We consider two types of MAX-SAT problems. The first formulation (\textit{forward SAT}) is used while the routes are being computed, in order to optimise the rollout outputs. The second solution (\textit{tightening SAT}) is used to tighten the final outputs after the episode is completed.

\subsubsection{Forward SAT}
In the following description, we may omit the vehicle subscript $k$ as the formulation optimises the proposed route for a specified vehicle only. Assume that a sub-tour $R$ is generated by the rollout policy as per Algorithm \ref{alg:pseudo}, consisting of a sequence $R_i\forall \text{ customers } c_i\in R$. We also define a set $A$ of additional customers which are active and which are cluster neighbours of the members of $R$.

We wish to compute optimised integer ordering variables $O_i$ and service start times $T_i$ for customers in $R\cup A$. The forward-SAT formulation is given by,
\begin{equation}
    J = \min_{O_i} \left[ \left(\sum_{i\in R\cup A} D_{i}\right) + D_{return}\right],\text{ where}
    \label{eq:objfwd}
\end{equation}
\begin{equation}
D_i = \begin{cases}
d_{loc,i} & \text{ if }O_i=1 \\
d_{j,i}   & \text{ if }O_i>1\text{ and }O_i=O_j+1\\
d_{max}   & \text{ if }O_i=-1\text{ i.e. }c_i\text{ not served}
\end{cases}
\end{equation}
\begin{equation}
D_{return}=d_{o,i}\text{ for last }c_i\text{ being served}
\end{equation}
The variable $D_i$ represents the leg distance to get to customer $c_i$. For the first customer on the sub-tour, the distance is computed from the current location $loc$ (see Table \ref{tab:inputs}). For all other legs, it is the distance between successive customer locations. We will assign a distance penalty $d_{max}$ as defined in Sec. \ref{subsec:preproc} for any customers that are skipped. Finally, $D_{return}$ is the distance to return to the depot, should the vehicle be unable to serve any further customers. Including this term encourages closed-loop journeys over ones which could end very far from the depot. 

Note that $J$ in (\ref{eq:objfwd}) is equivalent to the one defined in (\ref{eq:objective}), but is only reformulated in terms of ordering variables $O_i$. Similarly, we can easily formulate the system constraints (\ref{eq:tw}), (\ref{eq:maxload}), (\ref{eq:depottimestart}), and (\ref{eq:intercustomer}) with the understanding that $(f_{i}=1) 
\equiv (O_i=1)$ and $(a_{i,j}=1) \equiv (O_j=O_i+1)$.

As formulated above, the solver is able to generate solutions within 1 second for small problems (sub-tours consisting of up to 6 customers). To further improve the computational time, we trade off some optimality by introducing constrained position shifting \cite{balakrishnan2006scheduling}. The CPS constraints are given by,
\begin{equation}
(R_i- \Delta \leq O_i \leq R_i + \Delta)\,\,\forall\,c_i\in R
\label{eq:cps}
\end{equation}
\begin{equation}
O_i \geq 1\,\,\forall\,c_i\in R
\label{eq:forced}
\end{equation}
Constraint (\ref{eq:cps}) restricts the deviation in order for the customers that are already part of the sub-tour $R$. Note that the additional customers in $A$ are not included. The final constraint (\ref{eq:forced}) forces the solver to serve all customers already in $R$.

\subsubsection{Tightening SAT}
The MAX-SAT formulation described in (\ref{eq:objfwd})-(\ref{eq:forced}) can be reused for tightening the final routes after episode completion, with minor changes. The key change is that we only consider the tour $R$, without any additional opportunistic customers. We also start with a larger value of CPS parameter $\Delta$, to explore potentially greater optimisation opportunities. Finally, we note that the tour must start and end at the depot, which implies that $loc \equiv o$.

\section{Results and discussion} \label{sec:results}

\begin{figure}
\centering
\includegraphics[width=0.4\textwidth]{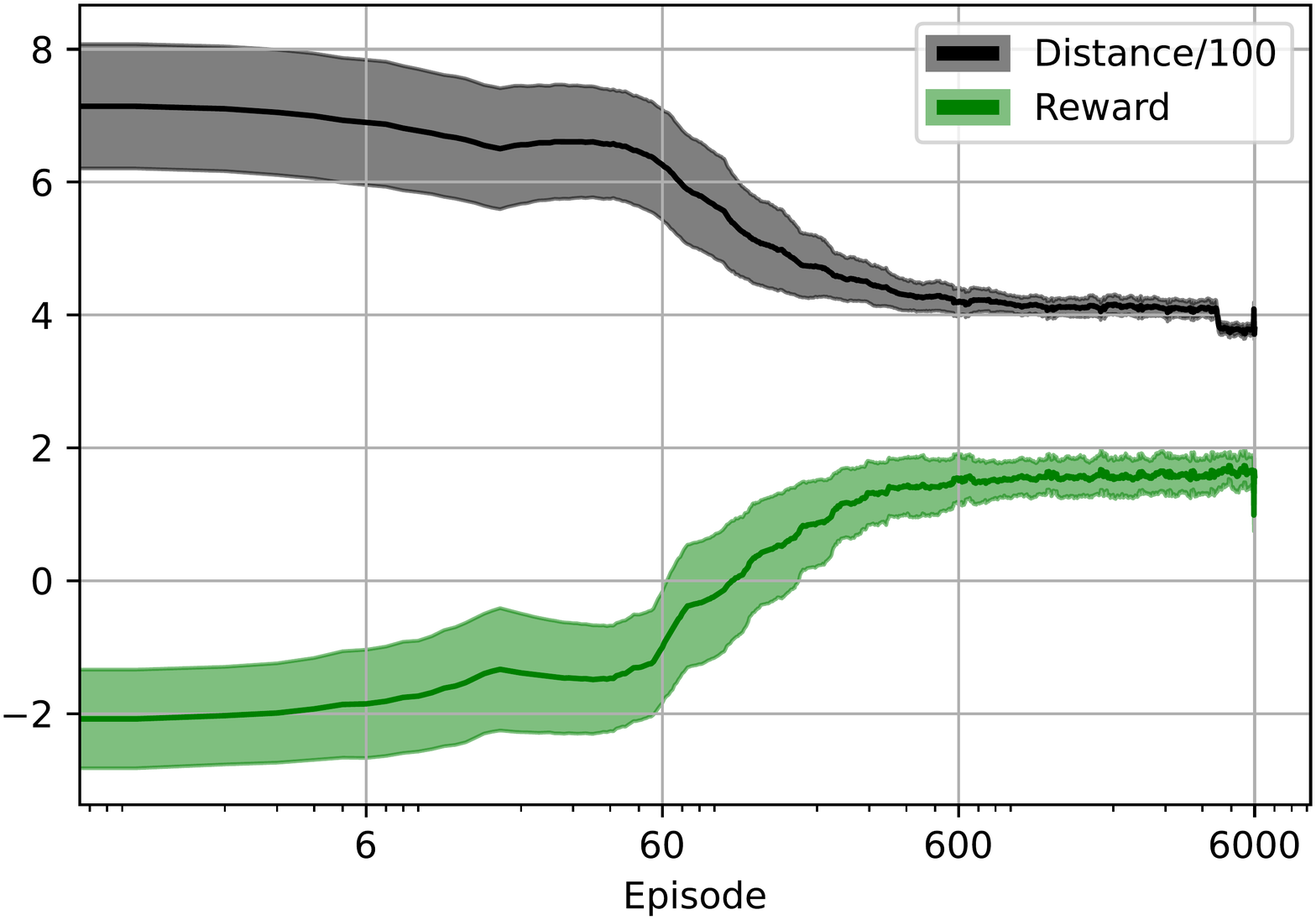}
\caption{Training results over three random seeds, using the Solomon 25-customer data sets. Note that the \textit{x}-axis is logarithmic.}
\label{fig:training}
\end{figure}

The RL algorithm described in Sec. \ref{subsec:rl} is trained on all instances of the Solomon data set \cite{solomondata} containing 25 customers (a total of 56 data sets). The results of training over 6000 episodes for 3 random seeds is shown in Fig. \ref{fig:training}, with the hyperparameters as described in Sec. \ref{subsec:rl}. The MAX-SAT formulation is only activated for the final 2000 episodes, with the first 4000 episodes proceeding with RL and rollouts only. A steady trend is seen in both the rewards received and the moving average of distance over episodes.

The testing of the learnt model is carried out on all 56 data sets in the Solomon data \cite{solomondata} for 25, 50, and 100 customers. There are three types of geographical distributions: C stands for clear clusters, R stands for a random distribution, while RC stands for a mix of clustered and random. Additionally, data sets tagged with `1' (as in RC1) have low vehicle capacity, while ones tagged with 2 have higher vehicle capacity. The first of these is the same set that is used for training, while the latter two are unseen. The results are compiled in Table \ref{tab:inputs}. The first column in the table is the average best-known distance in literature. The next two columns GA and RH are values reported in \cite{gupta2022deep}, while QL are the values reported in \cite{sultana2021fast}. The final two columns are from the current proposal, with RL+RO including only rollouts while RL+SAT+RO also includes the MAX-SAT formulation.

Table \ref{tab:inputs} also reports the average computation time at the start of each set of data, wherever available. A comparison of the distances and computation times shows a clear tradeoff between solution quality (lower distances are better) and computation times. Note that in a few instances, RL+SAT+RO outperforms GA on both route distance and computation time. In one type of distribution (C2), the proposed method appears to result in somewhat poor performance. A deeper look at the generated routes in this case reveals that the learnt RL policy does not foresee the possibility of a single vehicle with high capacity serving multiple clusters in the data. This is a discrepancy which will be addressed in future work.

\begin{figure}
\centering
\includegraphics[width=0.4\textwidth]{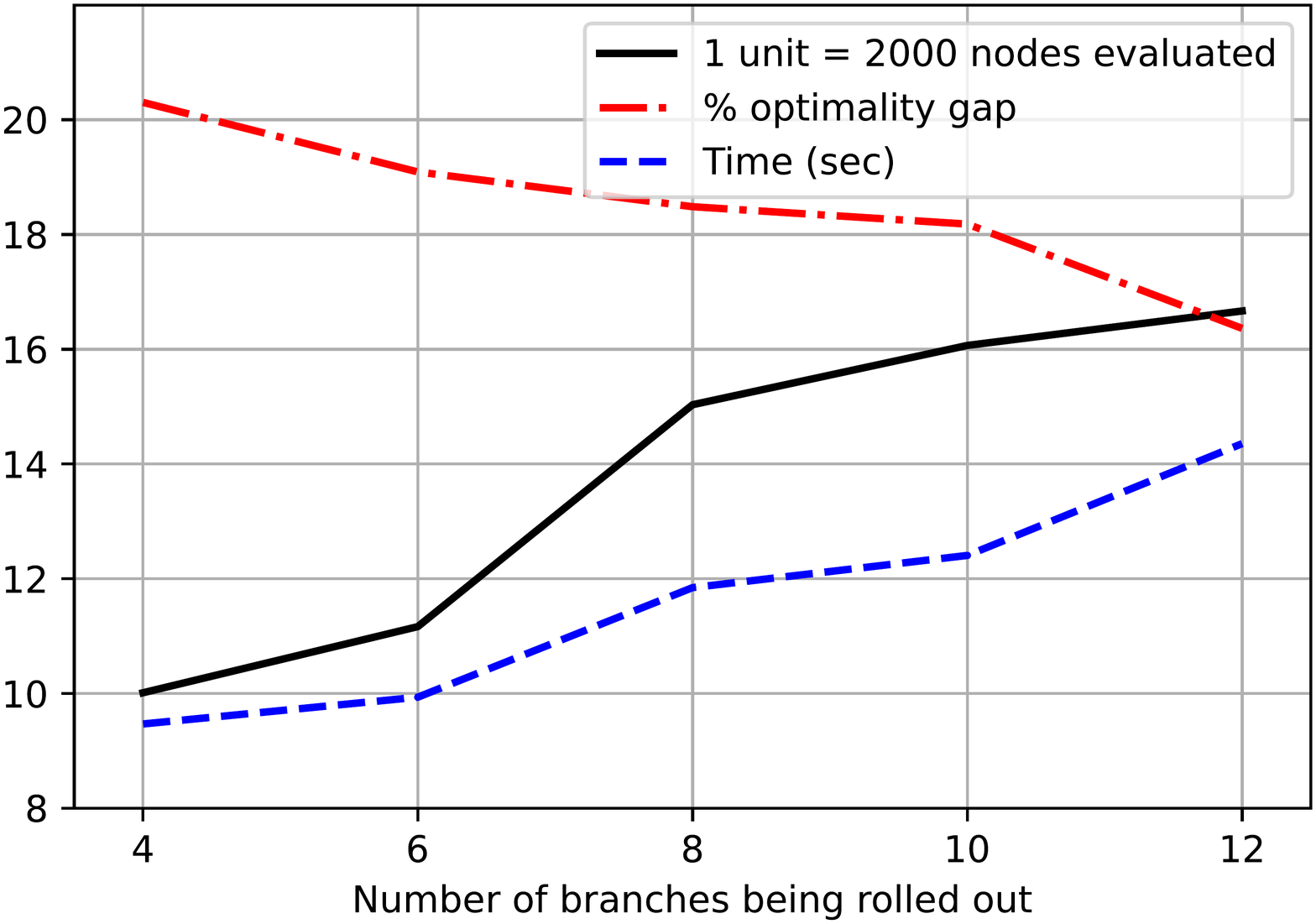}
\caption{Tradeoff between rollout extent, solution quality, computation time.}
\label{fig:variation}
\end{figure}

Finally, the tradeoff in computation time and solution quality is made clear by the experiment reported in Fig. \ref{fig:variation}, where we see the effect of varying $\kappa$ (number of rollout branches) on computation time, number of nodes evaluated, and the optimality gap of the resulting solution. We see that as the number of rollouts increase, the computation time increases steadily while the solution quality improves. This provides justification for the claim that the proposed RL+RO+SAT approach allows us to choose from a spectrum of policies from pure RL (towards lower value of $\kappa$) to full meta-heuristics (high values of $\kappa$, with multiple rollouts in each branch).

In conclusion, this paper bridges the gap between three extreme approaches for CVRP-TW, namely heuristics (high design time), meta-heuristisc (high computation time), and reinforcement learning (poor solution quality). Future work in this area will attempt to formalise the counterfactual objective of setting the optimal hyperparameters of RL+RO+SAT given a target for computation time and solution quality.

\begin{table}
\centering
\caption{Comparison of results, with numbers representing average distance of the routes in a given type of data. Approximate computation times are listed at the top of each column. Question marks indicate values which are not available in literature. The first column is the best known distance compiled from multiple sources \cite{solomondata}. Methods GA and RH are from \cite{gupta2022deep}, while QL is from \cite{sultana2021fast}. The final two columns are from the current approach, with $RL+RO$ using rollouts only, and $RL+SAT+RO$ also including the SAT solver.}
\label{tab:inputs}
\begin{tabular}{|p{1.1cm}d{4.0}d{4.0}d{4.0}d{4.0}d{4.0}d{4.0}|}
\hline
     &      &        &              &      &         & RL+    \\
Data & Best & GA & RH & QL & RL +    & SAT+ \\
     & dist &        &        &        & RO & RO \\
\hline
     & (8400s) & (35s) & (1s) & (5s) & (8s) & (15s) \\
C1-25   & 191 & 191 & 283 & 215 & 199 & \textbf{192} \\
R1-25   & 464 & 469 & 619 & 586 & 620 & \textbf{568} \\
RC1-25  & 350 & 351 & 380 & 489 & 395 & \textbf{362} \\
C2-25   & 216 & 245 & 297 & 246 & 273 & \textbf{261} \\
R2-25   & 382 & 420 & 540 & 521 & 563 & \textbf{513} \\
RC2-25  & 319 & 365 & 598 & 448 & 352 & \textbf{329} \\
\hline
     & (??) & (150s) & (2s) & (30s) & (25s) & (60s) \\
C1-50   & 362 & 362 & 676 & 407 & 398 & \textbf{374} \\
R1-50   & 766 & 785 & 1114 & 982 & 1067 & \textbf{963} \\
RC1-50  & 730 & 741 & 1011 & 951 & 963 & \textbf{849} \\
C2-50   & 357 & 401 & 775 & 457 & 569 & \textbf{525} \\
R2-50   & 634 & 652 & 989 & 891 & 878 & \textbf{824} \\
RC2-50  & 585 & 621 & 1203 & 811 & 673 & \textbf{631} \\
\hline
     & (??) & (1200s) & (??) & (90s) & (150s) & (350s) \\
C1-100  & 826 & 830 & (??) & 975 & 914 & \textbf{869} \\
R1-100  & 1210 & 1210 & (??) & 1620 & 1790 & \textbf{1646} \\
RC1-100 & 1384 & 1385 & (??) & 1800 & 1977 & \textbf{1788} \\
C2-100  & 587 & 589 & (??) & 787 & 1105 & \textbf{1012} \\
R2-100  & 902 & 902 & (??) & 1341 & 1428 & \textbf{1309} \\
RC2-100 & 1063 & 1063 & (??) & 1497 & 1548 & \textbf{1427} \\
\hline
\end{tabular}
\end{table}

\bibliographystyle{IEEEtran}
\bibliography{refs}

\end{document}